\documentclass[conference]{IEEEtran}
\usepackage{times}

\usepackage[square,numbers,sort,compress]{natbib}
\usepackage{multicol}
\usepackage[bookmarks=true]{hyperref}
\usepackage{booktabs}
\usepackage{graphicx}
\usepackage{cuted}
\usepackage{caption}
\usepackage{amsmath}
\usepackage{amsfonts}
\usepackage{xcolor}

\newcommand{\Enc}{E_{\phi}}
\newcommand{\Dec}{D_{\theta}}
\newcommand{\dataset}{\mathcal{D}}
\newcommand{\Dyn}{F_{\psi}}

\begin{document}

\title{Interactive World Simulator \\for Robot Policy Training and Evaluation}

\author{\authorblockN{Yixuan Wang$^{1}$\quad Rhythm Syed$^{1}$\quad Fangyu Wu$^{1}$\quad Mengchao Zhang$^{2}$\quad Aykut Onol$^{2}$\quad Jose Barreiros$^{2}$\\Hooshang Nayyeri$^{3}$\quad Tony Dear$^{1}$\quad Huan Zhang$^{4}$\quad Yunzhu Li$^1$}
\authorblockA{$^1$Columbia University\quad $^2$Toyota Research Institute \quad $^3$Amazon \quad $^4$University of Illinois Urbana-Champaign}
\authorblockA{\textbf{\textcolor{magenta}{\url{https://www.yixuanwang.me/interactive_world_sim/}}}}
}

\maketitle

\begin{strip}
    \centering
    \includegraphics[width=2.0\columnwidth]{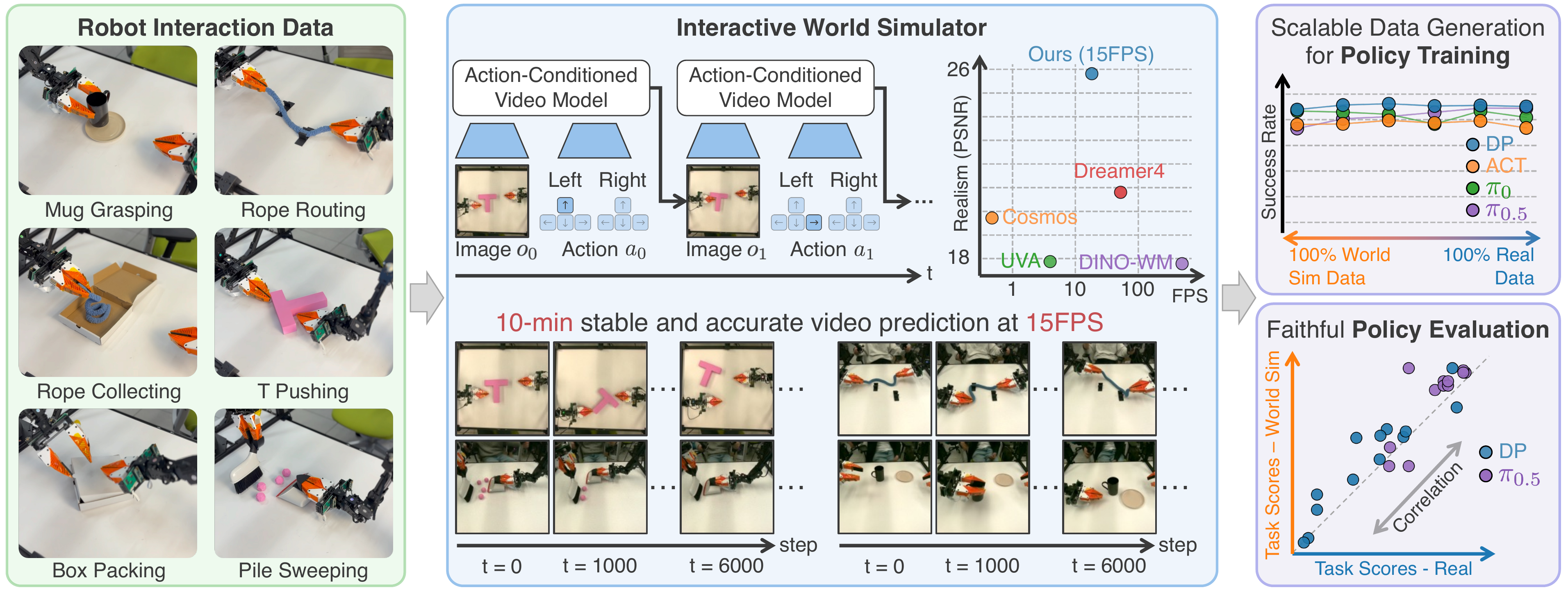}
    \captionof{figure}{\small \textbf{Overview of the Interactive World Simulator.}
    Given real-world robot interaction datasets (left), we train action-conditioned video models that capture complex physical dynamics and support stable long-horizon interactions (middle). The resulting world models achieve stable video prediction for over \textbf{10 minutes at 15 FPS on a single RTX 4090 GPU}, outperforming prior world models in long-horizon realism. The resulting Interactive World Simulator enables \textbf{scalable data generation for imitation policy training} without requiring additional robot interaction and supports \textbf{faithful and reproducible policy evaluation}, exhibiting a strong correlation between policy performance in simulation and in the real world (right). Additional examples are available on the \href{https://www.yixuanwang.me/interactive_world_sim/}{project website}.
    }
    \label{fig:teaser}
\end{strip}

\begin{abstract}

Action-conditioned video prediction models (often referred to as world models) have shown strong potential for robotics applications, but existing approaches are often slow and struggle to capture physically consistent interactions over long horizons, limiting their usefulness for scalable robot policy training and evaluation.
We present Interactive World Simulator, a framework for building interactive world models from a moderate-sized robot interaction dataset. Our approach leverages consistency models for both image decoding and latent-space dynamics prediction, enabling fast and stable simulation of physical interactions. In our experiments, the learned world models produce interaction-consistent pixel-level predictions and support stable long-horizon interactions for more than \textit{10 minutes at 15 FPS on a single RTX 4090 GPU}.
Our framework enables scalable demonstration collection solely within the world models to train state-of-the-art imitation policies. Through extensive real-world evaluation across diverse tasks involving rigid objects, deformable objects, object piles, and their interactions, we find that policies trained on world-model-generated data perform comparably to those trained on the same amount of real-world data.
Additionally, we evaluate policies both within the world models and in the real world across diverse tasks, and observe a strong correlation between simulated and real-world performance.
Together, these results establish the Interactive World Simulator as a stable and physically consistent surrogate for scalable robotic data generation and faithful, reproducible policy evaluation.

\end{abstract}

\IEEEpeerreviewmaketitle

\section{Introduction}

Video prediction models have shown promising results for robotic manipulation, including planning~\cite{chen2025large}, control~\cite{hafner2023mastering, zhou2024dino}, policy steering~\cite{wu2025foresight}, and policy evaluation~\cite{team2025evaluating}. However, existing action-conditioned video prediction models are either computationally expensive, due to heavy neural networks and multi-step diffusion processes~\cite{agarwal2025cosmos, guo2025ctrl, li2025unified}, or unstable over long-horizon rollouts due to accumulated prediction errors~\cite{zhou2024dino, hafner2023mastering}. As a result, faithful long-horizon action-conditioned video prediction of complex physical interactions remains challenging for state-of-the-art models, limiting their applicability to scalable policy training data generation and reproducible policy evaluation.

\begin{figure*}
    \centering
    \includegraphics[width=2.0\columnwidth]{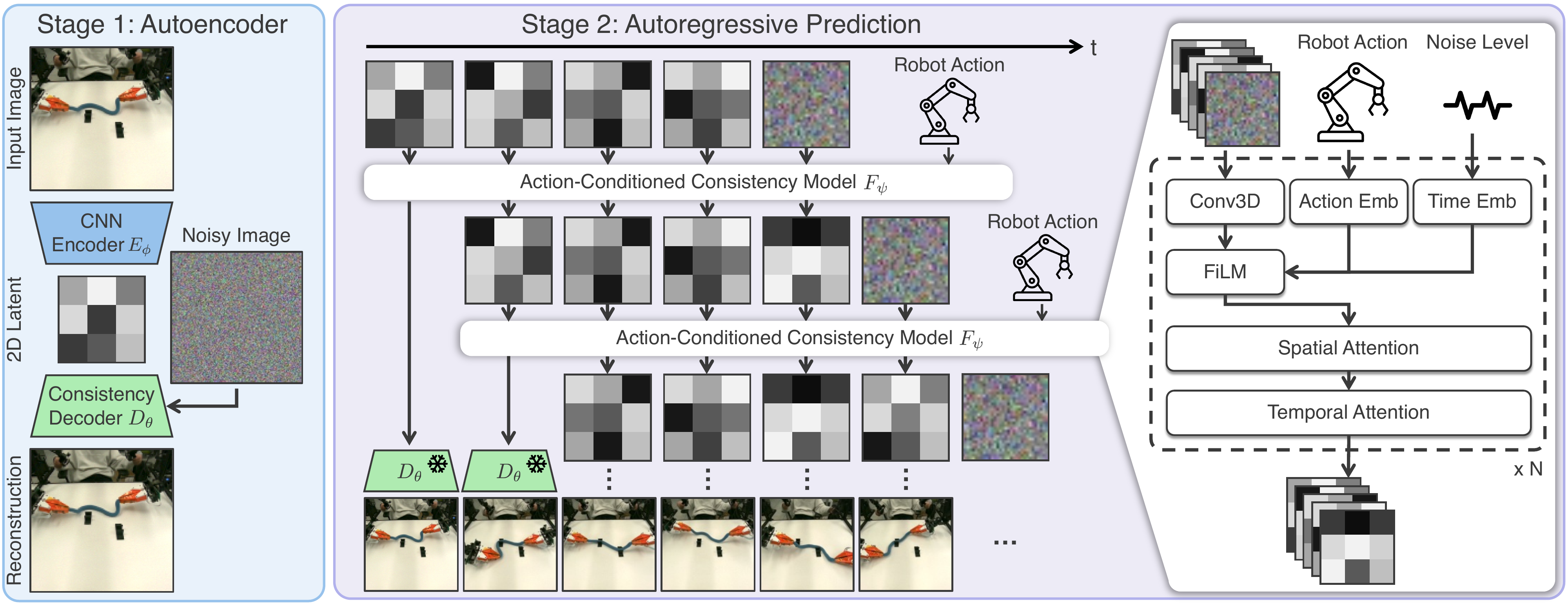}
    \caption{\small \textbf{Method Overview.} Model training proceeds in two stages. In Stage 1, we train an autoencoder that maps RGB observations into compact 2D latent representations using a CNN encoder $\Enc{}$ and a consistency-model decoder $\Dec{}$, enabling efficient and high-fidelity reconstruction. In Stage 2, we freeze the autoencoder and train an action-conditioned consistency model $\Dyn{}$ in latent space to model future visual dynamics using next-frame supervision. We choose $\Dyn{}$ as a consistency model because it efficiently represents multimodal distributions over possible future outcomes. The dynamics model learns to denoise noisy latent states conditioned on robot actions, past latents, and noise levels, and is instantiated as a stack of 3D convolutional blocks with FiLM modulation and spatiotemporal attention. During inference, $\Dyn{}$ autoregressively predicts future latents, which are decoded by $\Dec{}$ to generate long-horizon video predictions.}
    \label{fig:method}
\end{figure*}

In this work, we introduce \textit{Interactive World Simulator} (Figure~\ref{fig:teaser}), an action-conditioned video prediction model that supports stable, interactive rollouts for more than \textbf{10 minutes at 15 FPS on a single RTX 4090 GPU}. To achieve efficient long-horizon prediction, we first encode high-dimensional images into compact 2D latent representations and perform future prediction entirely in latent space. In the first stage, we train an autoencoder using a CNN encoder~\cite{lecun2002gradient} and a consistency-model decoder~\cite{song2023consistency} to enable efficient and high-fidelity reconstruction. In the second stage, we freeze the autoencoder and train an action-conditioned dynamics model using next-frame supervision in latent space. We model latent dynamics using consistency models, which are computationally efficient and capable of representing the multimodal distribution of possible future outcomes arising from robotic interaction. During inference, we autoregressively shift a fixed-length context window to generate long-horizon video predictions conditioned on robot actions, as shown in Figure~\ref{fig:method}.

Our Interactive World Simulator enables two important robotic applications:
\textbf{1) Scalable high-quality data generation for imitation policy training.}
Imitation learning has demonstrated strong performance in robotic manipulation~\cite{chi2025diffusion, zhao2023learning, black2024pi_0, intelligence2025pi_}, but existing approaches typically require large amounts of real-robot data, which are expensive to collect and difficult to scale. Our world simulator provides a realistic interactive environment in which users can collect demonstration data directly through interaction, enabling large-scale data collection without access to physical robots and substantially reducing data collection cost.
\textbf{2) Reproducible policy evaluation within the Interactive World Simulator.}
Policy evaluation is a longstanding challenge in robotics and is critical for algorithm iteration and checkpoint selection. As noted in~\cite{barreiros2025careful}, real-world evaluation is time-consuming and difficult to conduct in a controlled, apples-to-apples manner, which slows policy development and complicates fair comparison between methods. In contrast, policy evaluation within our world simulator is scalable and reproducible, enabling efficient and consistent comparison across policies. Because our model is trained on real-world robot interaction data, it exhibits reduced domain gap and dynamics mismatch compared to conventional simulators.

We conduct comprehensive experiments comparing our action-conditioned video prediction model with state-of-the-art baselines across a diverse set of tasks involving rigid objects, deformable objects, articulated objects, object piles, and their interactions, as shown in Figure~\ref{fig:teaser}. Our model outperforms prior methods, including Cosmos~\cite{agarwal2025cosmos}, UVA~\cite{li2025unified}, DINO-WM~\cite{zhou2024dino}, and Dreamer4~\cite{hafner2025training}, in terms of long-horizon prediction realism, while maintaining \textbf{interactive performance at 15 FPS on a single RTX 4090 GPU}.

To study whether our world simulator can generate data of comparable quality to real-world demonstrations, we collect the same amount of expert data within the simulator under identical initial configurations and data collection protocols. We then train imitation policies, including Diffusion Policy (DP), Action Chunking Transformer (ACT), $\pi_0$, and $\pi_{0.5}$~\cite{chi2025diffusion, zhao2023learning, black2024pi_0, intelligence2025pi_}, using different mixtures of real-world and simulator-generated data. Across all mixture ratios, ranging from 100\% world-simulator data to 100\% real-world data, we observe comparable policy performance, indicating that simulator-generated data are of similar quality to real-world data. Finally, we evaluate policies both in the real world and in the world simulator across multiple tasks and training checkpoints, and observe a strong correlation between simulator and real-world performance, demonstrating that our world simulator can faithfully reflect relative policy performance.

In summary, our main contributions are threefold: 1) we introduce the \textit{Interactive World Simulator}, an interactive action-conditioned video prediction model that supports stable long-horizon rollouts for over 10 minutes at 15 FPS across complicated physical interactions involving rigid objects, deformable objects, object piles, and multi-object interactions; 2) using this simulator, we enable scalable, high-quality data collection for imitation learning without requiring access to physical robots; and 3) we demonstrate through extensive experiments that policy performance in the world simulator strongly correlates with real-world performance, enabling reproducible and scalable policy evaluation.

\section{Related Works} 

\subsection{Video Prediction Model for Robotic Manipulation}

Video prediction models, or world models, have shown significant potential in robotics \cite{agarwal2025cosmos, alonso2024diffusion, ball2025genie, brooks2024video, liang2024dreamitate, liang2025video, kim2026cosmos, jang2025dreamgen, mei2026video, wiedemer2025video, escontrela2023video, fragkiadaki2015learning, agrawal2016learning, singh2025deep, allshire2025visual, du2023learning, wu2023unleashing, cheang2024gr2generativevideolanguageactionmodel, ye2024latentactionpretrainingvideos, wen2023any, ranasinghe2026future, gao2026dreamdojo, reuss2024multimodal, tian2024predictive, zheng2025flare, guo2024prediction, hu2024video, bu2025univla, chen2025unified, routray2025vipra, li2026causal}, serving critical roles in zero-shot planning~\cite{li2025novaflow,chen2025large}, policy steering \cite{wu2025foresight}, evaluation \cite{team2025evaluating}, and direct control. Frameworks like NovaFlow \cite{li2025novaflow} and RIGVid \cite{patel2025robotic} demonstrate the power of repurposing large-scale video generation to derive actionable 3D object flow and trajectories for manipulation without real-world demonstrations. Recent advancements such as the Large Video Planner \cite{chen2025large} further establish video as a primary modality for robot foundation models, enabling generative planning in pixel space. These models are also utilized for run-time policy steering. FOREWARN \cite{wu2025foresight} uses an action-conditioned world model alongside a Vision-Language model (VLM) to select optimal plans. Additionally, Ctrl-World \cite{guo2025ctrl} and VEO \cite{team2025evaluating} leverage large-capacity video models for evaluation such as Stable Video Diffusion (SVD) \cite{blattmann2023stable} and VEO2 \cite{vandenOord2024veo} respectively, providing high-fidelity simulation of nominal and out-of-distribution scenarios to assess safety and performance. 

However, existing video prediction models face significant practical limitations. Many state-of-the-art architectures, including closed-source models such as Sora \cite{brooks2024video} and open-source frameworks such as Diffusion Forcing \cite{chen2024diffusion} and the Diffusion Forcing Transformer (DFoT) \cite{song2025history}, are often not explicitly conditioned on robot actions, which is essential for grounding models in physical interaction. Furthermore, many high-capacity diffusion-based models~\cite{li2025unified, agarwal2025cosmos, guo2025ctrl, pai2025mimic, ye2026world} are not efficient enough for real-time interaction, often requiring enterprise-level GPU clusters that are inaccessible especially to many academic labs. Finally, some of existing models~\cite{zhou2024dino, hafner2023mastering, assran2025v} lack the robustness required for stable, long-horizon prediction. In contrast to the aforementioned methods, our Interactive World Simulator produces physically accurate pixel-level predictions, supports stable interactions for over 10 minutes at 15 FPS on a single consumer RTX 4090 GPU, making it highly accessible.

\subsection{Imitation Policy Training}

Imitation Learning (IL) has demonstrated remarkable results across a variety of robotic tasks through architectures such as Diffusion Policy \cite{chi2025diffusion}, Action Chunking Transformer (ACT) \cite{zhao2023learning}, $\pi_0$ \cite{black2024pi_0} and $\pi_{0.5}$ \cite{intelligence2025pi_} and many others \cite{zitkovich2023rt, barreiros2025careful, bjorck2025gr00t, lee2025molmoact, shukor2025smolvla, team2503gemini, zhang2025real}. Generalist models like Octo \cite{team2024octo} have pushed the boundaries of what IL can achieve in unstructured environments. Despite this progress, these methods remain dependent on high-quality real-robot expert data, which are scarce, expensive to collect, and difficult to scale without constant access to physical hardware. Our framework addresses this bottleneck by providing a system to build interactive world models using a play-data interaction dataset. We demonstrate that imitation policies trained on data generated using our simulator perform comparably to those trained on real-world data, enabling researchers and enthusiasts alike to iterate and scale their data collection strategies without solely relying on physical robots. 

\subsection{Imitation Policy Evaluation}

Reliable policy evaluation is a cornerstone of robotics research \cite{badithela2025reliable, atreya2025roboarena, li2024evaluating, zhou2025autoeval, yang2025robot, wang2025roboeval, kress2024robot, quevedo2025worldgym}, with real-world deployment serving as the gold standard for assessing performance. However, real-world evaluation is costly, time-consuming, and difficult to scale, making it impractical for frequent algorithm iteration, checkpoint selection, and fair comparison across methods. As a result, there has been growing interest in developing scalable evaluation frameworks whose outcomes correlate well with real-world policy performance. Prior work has explored both 3D- and 2D-based approaches to address this challenge. 3D-based evaluation methods, such as Physically Embodied Gaussian Splatting (PEGASUS) \cite{meyer2024pegasus}, integrate geometry and physics to create digital twins for real-time interaction. Similarly, Real-to-Sim Eval \cite{zhang2025real} focuses on structured representations for soft-body interactions, but often carries environmental assumptions such as tabletop configurations. Although 2D-based evaluation systems like VEO \cite{team2025evaluating} offer a more flexible approach, they are often closed-source and not readily available for broad academic use. The research community has also introduced various simulation benchmarks to enable systematic study. Traditional physics engines like MuJoCo \cite{todorov2012mujoco}, robosuite \cite{zhu2020robosuite}, and others \cite{xiang2020sapien, nvidia2024isaacsim} provide foundational environments for large-scale training and evaluation. We introduce a framework that requires only paired 2D RGB images and actions, making it highly transferable and scalable across diverse manipulation tasks. By ensuring a strong correlation between performance in our simulator and the real world, we provide an open-source, accessible solution for all labs to conduct reproducible and accurate policy evaluation.

\section{Method}
In this section, we first describe our problem formulation in Section~\ref{sec:prob}. We then elaborate on model training and inference in Section~\ref{sec:wm}. In Section~\ref{sec:data_gen} and Section~\ref{sec:policy_eval}, we discuss how to use this model for scalable data generation and reproducible policy evaluation.

\subsection{Problem Formulation}
\label{sec:prob}

We assume that the robotic interaction dataset $\dataset{}$ consists of multiple episodes, each of which has the form $\mathcal{E}=\{(o_0, a_0), (o_1, a_1), ..., (o_T, a_T)\}$, where $o_t\in \mathbb{R}^{3 \times H\times W}$ is an RGB image at time $t$ and $a_t$ is the action at time $t$. Our goal is to build an action-conditioned video prediction model $\hat{o}_t = f(o_{t-N:t-1}, a_{t-N:t-1})$ that takes in $N$ history observations and predicts the next RGB frame $\hat{o}_t$ by minimizing $||\hat{o}_{t} - o_{t}||_2$, the difference from the ground truth observation.

\subsection{Interactive World Simulator}
\label{sec:wm}

As shown in Figure~\ref{fig:method}, we train our action-conditioned video prediction model in two stages. In the first stage, we train an autoencoder that can encode an image $o$ into a latent representation $z$ and decode the latent representation to reconstruct the input image. In the second stage, we freeze the encoder and decoder and train the action-conditioned dynamics model using next-frame supervision in latent space. We elaborate on the different stages in the following section.

\subsubsection{Stage 1: Autoencoder Training}
\label{sec:stage_1}

Consistency models are widely used for image generation due to their efficiency and quality~\cite{song2023consistency, song2023improved, kim2023consistency}. Therefore, we instantiate the decoder as a consistency model for high-quality image generation.
Due to instability of 1-step consistency model training, we are inspired by Consistency Trajectory Model (CTM) for stable consistency model training~\cite{kim2023consistency}.

To train the autoencoder and find its parameters $(\phi,\theta)$, for each training image $o$, we first obtain the 2D latent representations $z=\Enc{}(o)\in \mathbb{R}^{C\times H'\times W'}$, where $C$ is the latent channel.
We then apply the same noise to $o$ at two different sampled scales $\sigma_t > \sigma_s \ge 0$.
\begin{equation}
x_{\sigma_t} = \mathcal{N}(o;\sigma_t), \qquad x_{\sigma_s} = \mathcal{N}(o;\sigma_s),
\end{equation}
where $\mathcal{N}(\cdot;\sigma)$ denotes the forward noising operator at noise scale $\sigma$.
The decoder is trained to map the higher-noise input to the lower-noise target conditioned on $z$:
\begin{equation}
\hat{x}_{\sigma_s} = \Dec{}\!\left(x_{\sigma_t}; \sigma_t,\sigma_s, z\right),
\end{equation}
by minimizing a weighted regression loss
\begin{equation}
\mathcal{L}_{\text{AE}}
=
\mathbb{E}_{o,\sigma_t>\sigma_s}\!\left[
w(\sigma_t)\,
\left\|
\hat{x}_{\sigma_s} - x_{\sigma_s}
\right\|_2^2
\right],
\end{equation}
where $w(\sigma)$ is a noise-dependent weight to balance learning across noise scales.
Thus, we obtain an image-to-latent encoder $\Enc{}(o)$ and a conditional generative decoder $\Dec{}(\cdot;\,z)$ that can reconstruct high-fidelity images from noisy inputs with a small number of denoising steps.

\subsubsection{Stage 2: Dynamics Training}
\label{sec:stage_2}

After stage~1, we freeze the autoencoder parameters $(\phi,\theta)$ and encode each frame $o_t$ into a latent $z_t$.
Our goal in stage~2 is to learn an action-conditioned latent dynamics model $\Dyn{}$ that predicts future latent frames given a context window of past latents $z_{t-N:t-1}$ and actions $a_{t-N:t-1}$.
Since consistency models can naturally model the multimodal distribution of potential futures while being efficient, we also model $\Dyn{}$ as a consistency model.

Concretely, we treat the latent sequence as a spatiotemporal tensor $Z \in \mathbb{R}^{C \times T \times H' \times W'}$.
Similar to stage 1, we apply the same noise to $Z$ using two different sampled scales $\sigma_s$ and $\sigma_t$, where $\sigma_t > \sigma_s \ge 0$.
But the major difference from stage 1 is that we only apply full noise to the last frame so that the dynamics model $\Dyn{}$ learns to predict the lower-noise last frame's latent given the action sequence and history context:
\begin{equation}
    \widehat{Z}_{\sigma_s}
    =
    \Dyn\!\left(Z_{\sigma_t};\, \sigma_t, \sigma_s, a_{t-N:t-1}\right).
\end{equation}
Figure~\ref{fig:method} shows how we appended the noisy latent to history latents during inference time.

We train $\Dyn$ with a weighted regression loss in latent space:
\begin{equation}
    \mathcal{L}_{\mathrm{dyn}}(\psi)
    =
    \mathbb{E}_{Z,\,\sigma_t>\sigma_s}\!\left[
    w(\sigma_t)\,
    \left\|
    \widehat{Z}_{\sigma_s} - Z_{\sigma_s}
    \right\|_2^2
    \right],
\end{equation}

We instantiate $\Dyn$ as a stack of 3D convolutional~\cite{tran2015learning} blocks with FiLM modulation~\cite{perez2018film} and spatiotemporal attention to fully capture spatial temporal relations.

To enable long-horizon action-conditioned prediction, one important technique is injecting small noise to observation contexts so that the dynamics model $\Dyn{}$ is robust to noisy contexts. During the online inference, models' prediction will be used as context for later steps, which is inevitably noisy. Therefore, model's robustness towards noisy context is very essential for stable long-horizon prediction.

\subsubsection{Inference}
\label{sec:inference}

During inference time, we predict future frames autoregressively. Given initial image $o_0$, we first obtain 2D latent $z_0=\Enc{}(o_0)$. As illustrated by Figure~\ref{fig:method}, we attach a noisy latent to history latents. Alongside action information, we denoise the last-frame noisy latent into a clean one (i.e., $z_1$). Then we attach the latest predicted latent to existing context to get the new history context (e.g., $z_{0:1}$). By repeating this process, we are able to autoregressively predict latents. Note that we discard the old latent once the context length is larger than a threshold so that the computation cost will not increase as horizon increases. Finally, the decoder $\Dec{}$ takes in newly predicted latents $z$ and renders new images $\hat{o}$.

\subsection{Data Generation for Policy Training}
\label{sec:data_gen}

Our Interactive World Simulator serves as a scalable surrogate for expert demonstration data collection, enabling the generation of high-quality synthetic demonstrations for imitation policy training without requiring access to physical robots.

Data collection is performed by initializing the simulator with an initial observation $o_0$, after which a human operator interacts with the simulator by issuing control commands through a teleoperation interface (e.g., keyboard input or low-cost kinematic devices). Given an action sequence $a_{0:T}$, the simulator autoregressively generates the corresponding observation sequence $o_{1:T}$, producing full demonstration trajectories $\{(o_t, a_t)\}_{t=0}^T$ that mirror real robot interaction data. The generated trajectories are directly compatible with standard imitation learning pipelines, including DP, ACT, and the $\pi$-series models, without requiring any modification to policy architectures or training procedures. This allows world simulator-generated data to be seamlessly mixed with real-world demonstrations or used as a standalone training source.

\begin{figure*}
    \centering
    \includegraphics[width=\linewidth]{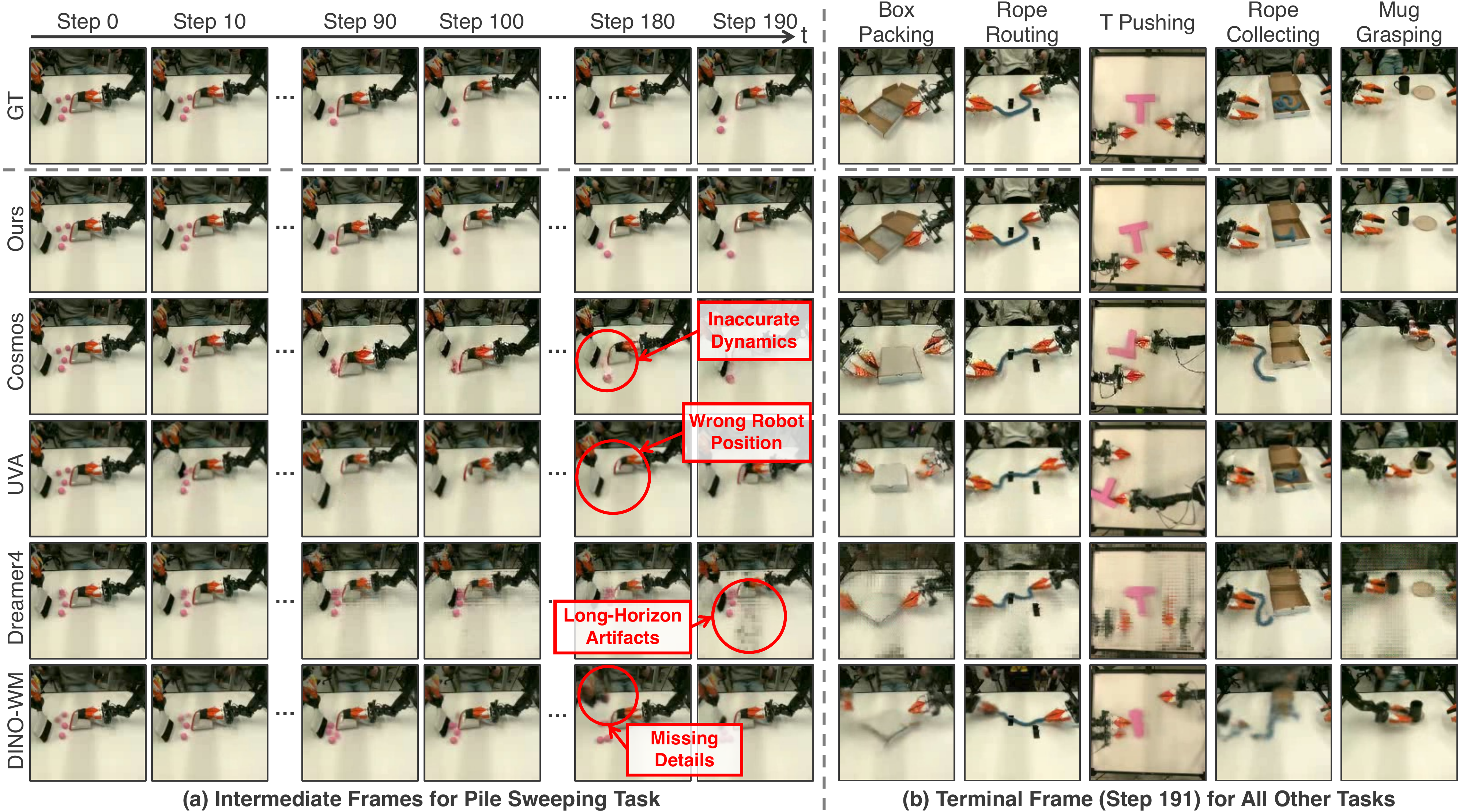}
    \caption{\small \textbf{Qualitative Comparison of Long-Horizon Action-Conditioned Video Prediction.} We compare the proposed Interactive World Simulator with state-of-the-art action-conditioned video prediction models across multiple manipulation tasks. (a) Intermediate predicted frames for the \texttt{Pile Sweeping} task illustrate long-horizon rollout behavior, where baseline methods exhibit inaccurate dynamics, robot pose drift, accumulated artifacts, or loss of fine-grained details over time. (b) Terminal predicted frames for additional tasks, including \texttt{Box Packing}, \texttt{Rope Routing}, \texttt{T Pushing}, \texttt{Rope Collecting}, and \texttt{Mug Grasping}, highlight differences in visual fidelity and interaction consistency. In contrast to baseline methods, our model preserves coherent robot–object interactions and maintains stable predictions over extended horizons.}
    \label{fig:quali}
\end{figure*}

\subsection{Policy Evaluation}

Evaluating policies fairly is another longstanding challenge in robotics. Performing real-world evaluation is time-consuming and difficult to scale, as each trial requires manual resets and careful control of experimental conditions such as the initial object configuration. Our Interactive World Simulator enables scalable policy evaluation by allowing policies to interact directly with the simulated environment: the world model takes policy actions and predicts future frames, while the policy consumes the predicted frames to generate subsequent actions, mirroring real-world interaction. Compared to physical experiments, interacting within the world simulator enables controllable initial configurations, faster experiment resets, and safer large-scale evaluation. In the experiments section, we study whether policy performance in the world simulator faithfully predicts performance in the real world.

\label{sec:policy_eval}

\section{Experiment}

In this section, we aim to study three questions: 1) How does our Interactive World Simulator perform compared to state-of-the-art world models in terms of realism, speed, and robustness? (Section~\ref{sec:vid_comp}) 2) Can our Interactive World Simulator generate data with quality similar to real data for imitation learning? (Section~\ref{sec:exp_data_gen}) 3) Can our Interactive World Simulator's policy evaluation faithfully represent real-world policy performance? (Section~\ref{sec:correlation})

\subsection{World Model Instantiation}
\label{sec:setup}
We set up one simulation task within MuJoCo~\cite{todorov2012mujoco} and six real-world tasks,
including \texttt{Mug Grasping}, \texttt{Rope Routing}, \texttt{Rope Collecting}, \texttt{T Pushing}, \texttt{Box Packing}, and \texttt{Pile Sweeping}, which involve rigid objects, deformable objects, object piles, articulated objects, and the interactions among them. We performed all tasks using the ALOHA Bimanual Robot \cite{zhao2023learning}.
In simulation, we instantiate the T pushing task similarly to its real-world counterpart, as shown in Figure~\ref{fig:teaser}. We use a scripted policy to generate 10,000 episodes of random interactions to ensure sufficient data coverage. In the real world, we collect about 600 episodes of play data for each task, with each episode consisting of 200 steps. Real-world data collection for each task takes around six hours for a single person, which is a manageable scale for most research labs.

To make tasks interactive, we constrain part of the tasks' action space. For example, we limit motion of robot effector to bimanual plane motion for T pushing task. Additionally, our default image resolution is $128\times 128$, which is sufficient for most of tasks.

Our model is lightweight. For example, the model for mug grasping task's size is 176.02 MB, making it easy to train and infer and keeps it very accessible for the broader research community. For stage 1 training, it usually takes around 6 hours on one H200 GPU, and around 12 hours for stage 2 training on one H200 GPU.

\begin{table*}[!ht]
\centering
\small
\begin{tabular}{lccccc}
\toprule
Metric & Ours & DINO-WM~\cite{zhou2024dino} & UVA~\cite{li2025unified} & Dreamer4~\cite{hafner2025training} & Cosmos~\cite{agarwal2025cosmos} \\
\midrule
MSE $\downarrow$ & \textbf{0.005} $\pm$ 0.005 & 0.028 $\pm$ 0.032 & 0.023 $\pm$ 0.015 & 0.012 $\pm$ 0.009 & 0.019 $\pm$ 0.010 \\
LPIPS $\downarrow$ & \textbf{0.051} $\pm$ 0.019 & 0.270 $\pm$ 0.093 & 0.272 $\pm$ 0.077 & 0.163 $\pm$ 0.052 & 0.224 $\pm$ 0.060 \\
FID $\downarrow$ & \textbf{63.50} $\pm$ 13.78 & 200.77 $\pm$ 79.02 & 142.55 $\pm$ 49.43 & 239.97 $\pm$ 45.75 & 200.74 $\pm$ 31.53 \\
PSNR $\uparrow$ & \textbf{25.82} $\pm$ 2.72 & 17.79 $\pm$ 2.84 & 17.87 $\pm$ 2.21 & 20.81 $\pm$ 2.21 & 18.91 $\pm$ 1.73 \\
SSIM $\uparrow$ & \textbf{0.831} $\pm$ 0.039 & 0.652 $\pm$ 0.059 & 0.650 $\pm$ 0.059 & 0.693 $\pm$ 0.045 & 0.647 $\pm$ 0.040 \\
UIQI $\uparrow$ & \textbf{0.960} $\pm$ 0.019 & 0.875 $\pm$ 0.029 & 0.884 $\pm$ 0.025 & 0.919 $\pm$ 0.024 & 0.883 $\pm$ 0.029 \\
FVD $\downarrow$ & \textbf{243.20} $\pm$ 103.58 & 1752.57 $\pm$ 805.56 & 2213.29 $\pm$ 525.48 & 1747.26 $\pm$ 248.08 & 799.34 $\pm$ 220.07 \\
\bottomrule
\end{tabular}
\caption{\small \textbf{Quantitative Comparison of Video Prediction Performance.} We quantitatively evaluate state-of-the-art action-conditioned video prediction models across a diverse set of manipulation tasks involving rigid objects, deformable objects, articulated objects, and object piles. The table reports results aggregated across all tasks; per-task results are provided in the supplementary material. Performance is measured using standard video prediction metrics, including MSE, LPIPS, FID, PSNR, SSIM, UIQI, and FVD, computed between predicted and ground-truth video rollouts. All results are reported for action-conditioned predictions over extended horizons (192 steps). The proposed Interactive World Simulator consistently outperforms prior world models across metrics, indicating improved visual fidelity, interaction consistency, and long-horizon stability, while maintaining efficient inference speed.}
\label{tab:video_comparison}
\end{table*}

\subsection{Video Baseline Comparisons}
\label{sec:vid_comp}
To evaluate the video prediction performance of the proposed Interactive World Simulator against prior action-conditioned world models, we conduct long-horizon action-conditioned video prediction comparisons across all six real tasks and one simulation task listed in Section~\ref{sec:setup}. We compare against representative state-of-the-art baselines, including Cosmos~\cite{agarwal2025cosmos}, UVA~\cite{li2025unified}, Dreamer4~\cite{hafner2025training}, and DINO-WM~\cite{zhou2024dino}, under identical initial conditions and rollout horizons (192 steps or 19.2 seconds). Prediction quality is assessed using standard reconstruction, perceptual, and temporal metrics such as MSE, PSNR, and FVD, computed between predicted and ground-truth video rollouts.

For baseline training, we follow the default settings in DINO-WM and Dreamer4 to train from scratch. We chose the community implementation version of Dreamer4 and followed their default training parameters. We followed the UVA training scripts to start the training from a pretrained VAE and another pretrained MAR model~\cite{li2024autoregressive}. Cosmos provides guidelines on finetuning the Cosmos model. Therefore, we converted our data to the Cosmos data format and followed the default Cosmos finetuning settings.

Quantitative results in Table~\ref{tab:video_comparison} show that the proposed method consistently outperforms prior methods across metrics, indicating improved visual fidelity and long-horizon temporal consistency. Table~\ref{tab:video_comparison} aggregates results across all tasks, and the full result for each task is attached in the appendix.

Qualitative comparisons in Figure~\ref{fig:quali} further reveal different failure patterns of baseline methods such as robot pose drift, inaccurate object dynamics, missing details, or severe artifacts over long rollouts, whereas the proposed model maintains physically plausible interactions and stable predictions. In addition, the Interactive World Simulator achieves significantly higher inference efficiency, running at up to 15 FPS on a single GPU, enabling interactive long-horizon prediction that is impractical for many diffusion-based baselines. Our model could also infer stably for more than 10 minutes at 15 FPS, and more videos can be found on our project website.

\begin{figure}[!ht]
    \centering
    \includegraphics[width=1.0\linewidth]{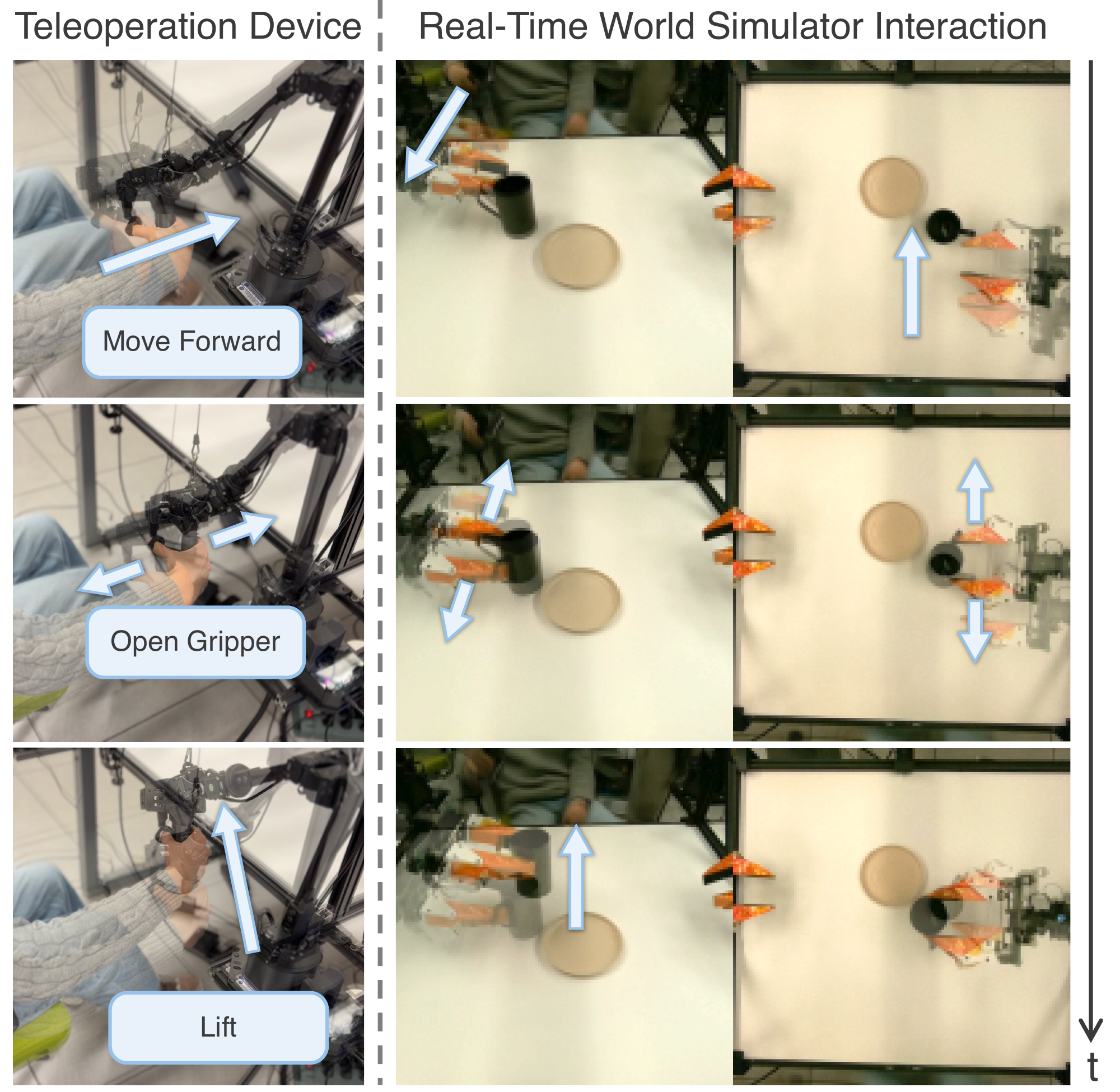}
    \caption{\small \textbf{Teleoperation Interface for Real-Time Interaction with the World Simulator.} We build a kinematic teleoperation device that allows users to control robot actions through the Interactive World Simulator. The left column shows a user issuing commands through the teleoperation device (e.g., moving forward, opening the gripper, and lifting the end-effector). The right columns show the corresponding frames generated by the world simulator in real time. The simulator follows the teleoperation commands and produces consistent robot–object interactions.}
    \label{fig:teleop}
    \vspace{-10pt}
\end{figure}

\begin{figure*}
    \centering
    \includegraphics[width=1.0\linewidth]{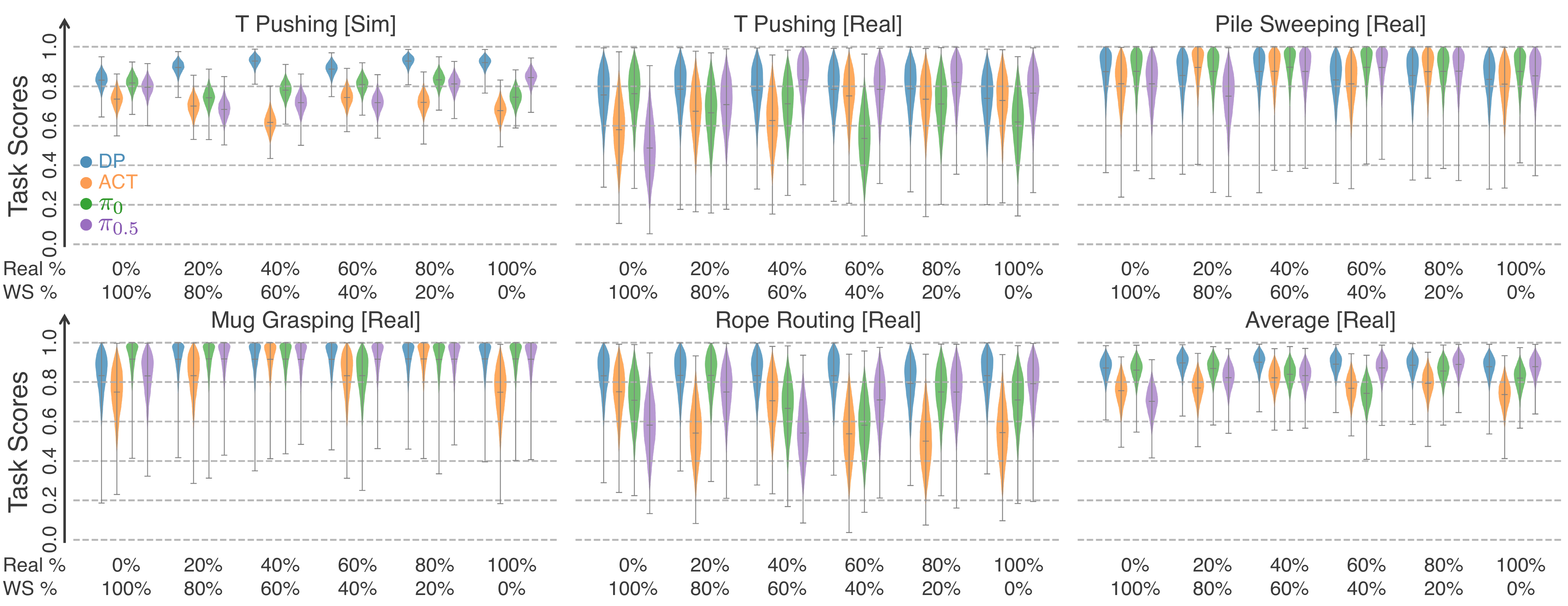}
    \caption{\small \textbf{Imitation Policy Training with Mixtures of Real-World Data and Our World Simulator Data.} ``WS \%'' represents the percentage of data from our world simulator, and ``Real \%'' represents the percentage of data from the real world. From left to right, the mixtures range from 100\% world simulator data to 100\% real-world data. We use a total of 100 training episodes for each policy and data mixture. Performance among the policies trained on different data mixtures across different manipulation tasks is observed to be consistently high with comparable task scores, suggesting that data generated by our world simulator is comparable in quality to real-world demonstrations.}
    \label{fig:data_gen}
\end{figure*}

\subsection{Data Generation for Policy Training}
\label{sec:exp_data_gen}

\begin{figure}[!ht]
    \centering
    \includegraphics[width=1.0\linewidth]{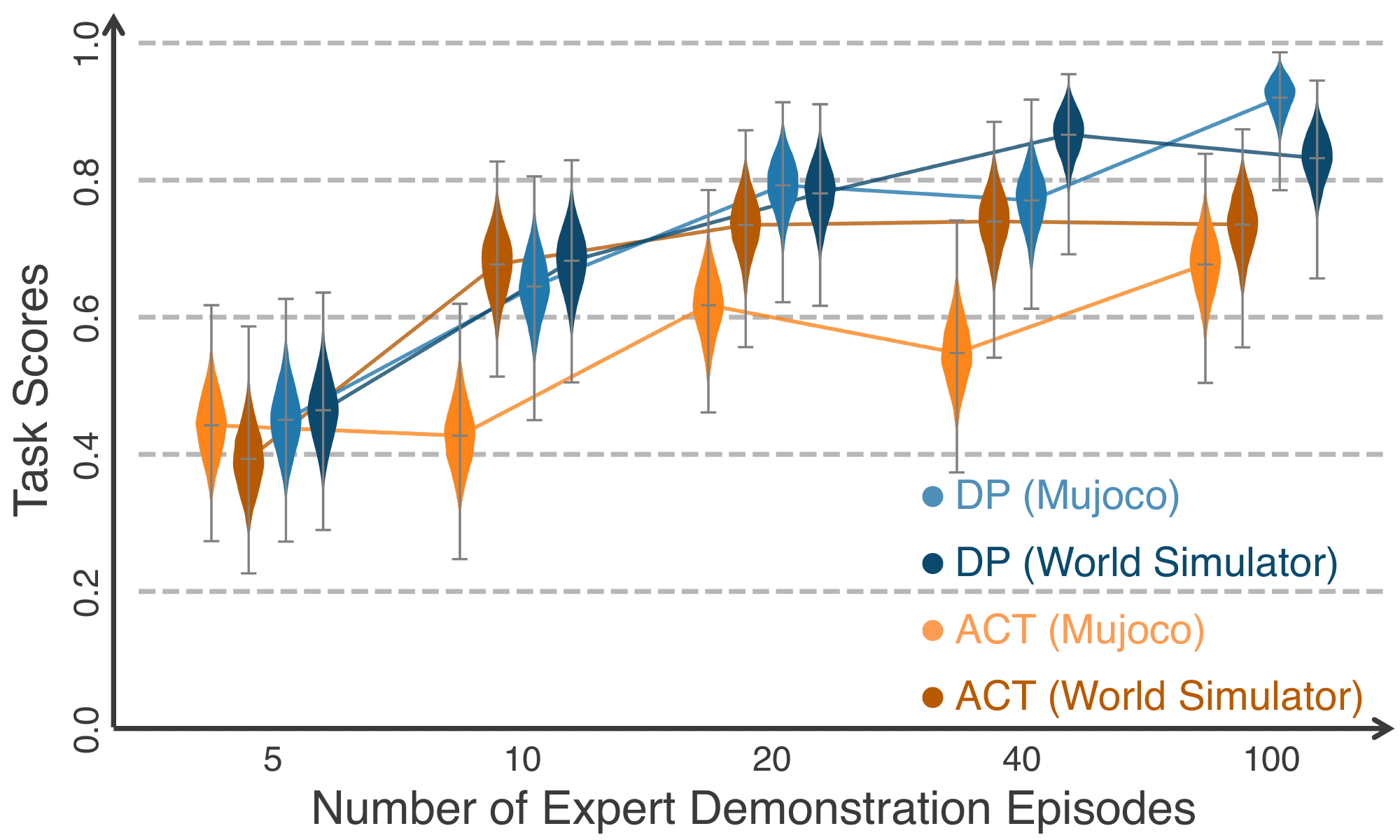}
    \caption{\small \textbf{Effect of Data Scaling on Policy Performance}. 
    We compare the performance of ACT and DP trained on datasets of varying sizes, ranging from 5 to 100 expert demonstration episodes. For each dataset size, policies are trained using either 100\% MuJoCo data or 100\% data generated from our world simulator. Policy performance improves consistently as the number of demonstrations increases across both data sources, suggesting that the scaling behavior of policy training is similar when using world simulator data.} \label{fig:data_mag_vs_score}
\end{figure}

To evaluate whether our world simulator can serve as a scalable data generator, we investigate whether synthetic demonstrations provide similar quality to real-world robot data for training imitation policies. We consider five manipulation tasks: T pushing in MuJoCo, T pushing in the real world, and three additional real-world tasks, pile sweeping, mug grasping, and rope routing. We benchmark four imitation learning policies: DP, ACT, $\pi_0$, and $\pi_{0.5}$. For each task, we use exactly 100 demonstration episodes to train imitation policies, regardless of the data source.

For fair and consistent policy evaluation, we define the task score for each task as follows across the whole paper, including Section~\ref{sec:correlation}:
(1)~T pushing: maximum intersection-over-union between the T block's current pose and its target pose achieved within 600 steps;
(2)~Rope routing: number of clips through which the rope is inserted within 200 steps;
(3)~Mug grasping: one point for grasping the mug and one point for placing it on the plate within 200 steps;
(4)~Pile sweeping: number of pile pieces swept into the tray within 200 steps. The rollout horizons for each task were chosen based on the average length of the expert demonstrations collected for the task plus some extra buffer of 100 steps to allow the policy rollout additional time to converge.

The user could use keyboard or kinematic devices to interact with the world simulator to collect expert demonstration data. Figure~\ref{fig:teleop} showcases the teleoperation experiences using kinematic devices. On the left, it shows the process of a user controlling the teleoperation device and pretending to grasp the mug. The right process happens entirely in the world simulator and is very close to the real-world rendering. But in fact, there are no mugs or plates on the table. This example showcases the immersive experience of the world simulator interaction, allowing the user to collect high-quality expert demonstration data for training imitation learning policies.

To evaluate the quality of the simulator's data, we conduct two experiments. First, we construct training sets using varying mixtures of real-world data and world simulator-generated data with constant 100-episode dataset sizes. Then we train imitation learning policies with these data mixture ratios, ranging from 100\% world simulator data to 100\% real data. For each data point in MuJoCo, we evaluate the policy 100 times. For each data point in the real world, we evaluate the policy 10 times. Then, we obtain Figure~\ref{fig:data_gen}, which represents task scores' Bayesian posteriors under a uniform Beta prior and the task scores we observe, following the practice of Large Behavior Models (LBM) analysis~\cite{barreiros2025careful}.

As shown in Figure~\ref{fig:data_gen}, we observe consistent policy performance across the entire spectrum of data mixtures. Our analysis confirms that policies trained on 100\% world simulator data achieve performance levels comparable to those trained on 100\% real world expert data. For DP, the average task scores remain extremely stable, with a score of $87.9\%$ using 100\% simulator data compared to $90.3\%$ using 100\% real data. Similarly, ACT demonstrates strong parity, achieving $76.2\%$ using simulator data versus $73.6\%$ using real data. While $\pi_{0.5}$ shows a general trend of improvement with increasing real data, rising from $73.1\%$ to $88.8\%$, it still achieves robust performance among the variety of tasks and initial configurations evaluated across simulation and real environments. Notably, policies trained on 100\% world simulator data perform comparably to those trained on an equivalent volume of real-robot expert data. This suggests that our simulator can generate data with quality similar to that of real-world demonstrations.

In the second set of experiments, we investigate whether the performance of imitation learning policies improves at a similar rate as additional data are introduced from different domains (real world vs. world simulator). We evaluate ACT and DP trained on datasets of varying sizes, ranging from 5 to 100 episodes. For each dataset size, we construct two training sets: one entirely from MuJoCo and one entirely from our world simulator. All policies are evaluated over 100 random trials. As shown in Figure~\ref{fig:data_mag_vs_score}, both policies exhibit consistent performance improvements as the training set size increases. Notably, data generated by our world simulator yield performance competitive with the MuJoCo baselines across all dataset sizes. In the low-data regime, policies trained on world simulator data successfully capture the task structure, matching the success rate of MuJoCo-trained policies. As the dataset scales to 100 episodes, the performance remains comparable. These results suggest that the scaling behavior of robotic policy training holds similarly within our world simulator, validating it as an effective tool for scalable data generation.

\begin{figure}[!ht]
    \centering
    \vspace{-5pt}
    \includegraphics[width=0.95\linewidth]{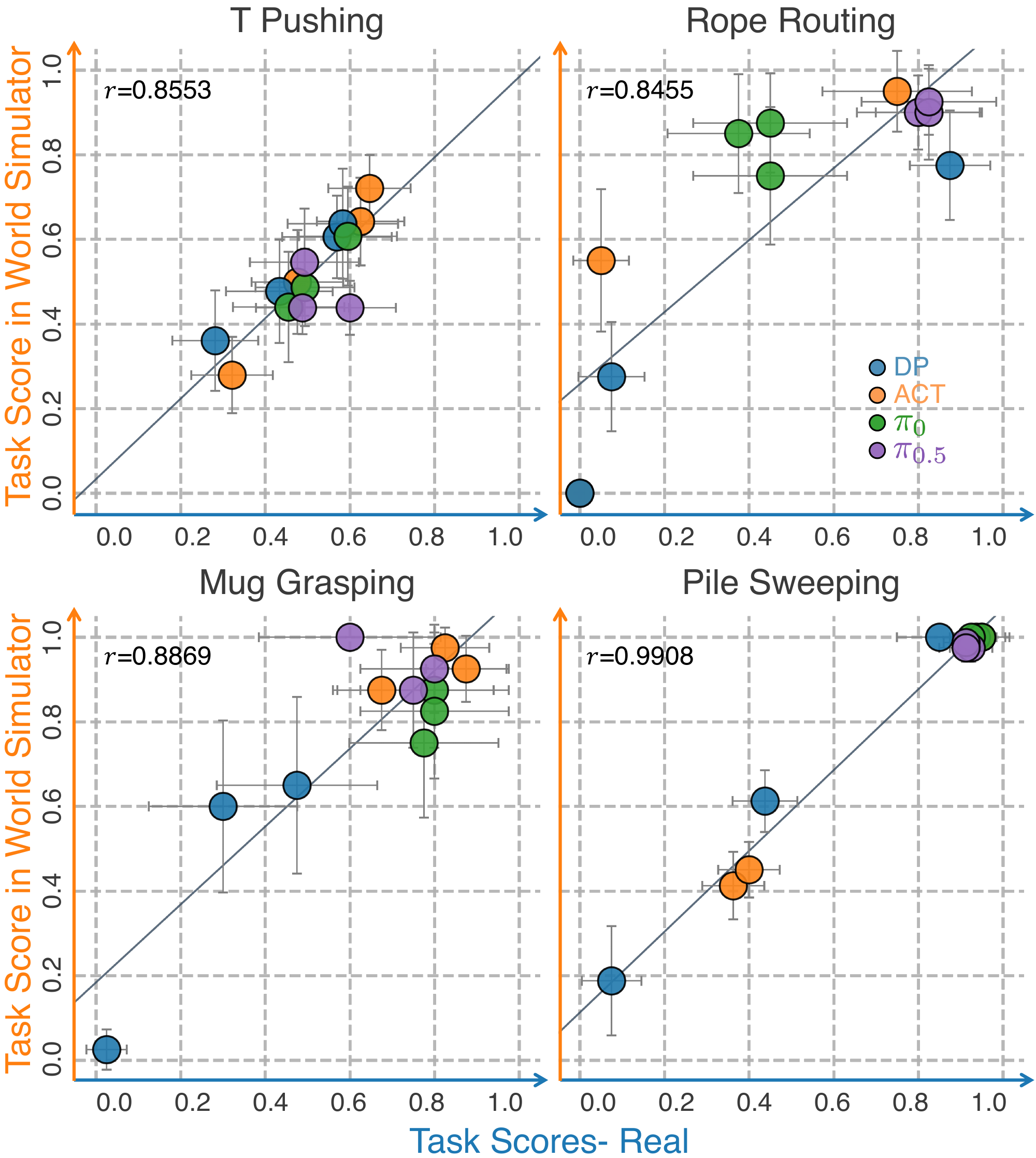}
    \caption{\small \textbf{Correlation Between Policy Performance in Our World Simulator and the Real World.} Real-world task performance of imitation policies is plotted against their corresponding performance in our world simulator across multiple manipulation tasks. Each point corresponds to a trained policy evaluated under identical settings in both environments. Strong positive correlations are observed, indicating that evaluation within our world simulator closely reflects relative policy performance in the real world.}
    \label{fig:correlation}
\end{figure}
\vspace{-5pt}

\subsection{Sim-to-Real Correlation for Faithful Policy Evaluation}
\label{sec:correlation}

We investigate whether our world simulator can serve as a proxy for real-world policy evaluation and faithfully represent real-world performance. For this to hold, policy performance in the world simulator must be positively correlated with performance in the real world under identical initial conditions.

We train four imitation policies, DP, ACT, $\pi_{0}$, and $\pi_{0.5}$, on real-world data and evaluate the final and intermediate checkpoints of the policies on four manipulation tasks: T pushing, rope routing, mug grasping, and pile sweeping.
For each task, we sample 20 initial configurations from the world simulator's training distribution and deploy each policy in both the world simulator and the real world.

Figure~\ref{fig:correlation} shows the normalized scores averaged over the 20 configurations, with the $x$-axis representing real-world performance and the $y$-axis representing the world simulator performance.
The error bars indicate Clopper-Pearson confidence intervals.
Across all four tasks, we observe strong positive correlations between simulator and real-world performance.
For tasks other than T pushing, the fitted lines exhibit a slight positive bias, indicating that policies tend to achieve slightly higher scores in the world simulator than in the real world.
Despite the sim-to-real gap, this bias does not diminish the simulator's utility: if one policy substantially outperforms another in the world simulator, our results suggest that it is likely to do so in the real world as well.
This property is useful in practice: one can rely on the world simulator to select high-quality candidate policies without the need for real-world experiments, thus reducing development cost and iteration cycles.

\section{Conclusion}

While action-conditioned video prediction shows great potential for robotics, existing approaches are often either too slow for interactive use or brittle under long-horizon inference. In this work, we introduced the Interactive World Simulator, an action-conditioned video world model that supports long-horizon, stable, and physically consistent robot interactions. The simulator enables efficient video prediction for over 10 minutes of continuous interaction at 15 FPS while maintaining high-quality pixel-level predictions. Our approach addresses key limitations of prior world models for robotics, improving both stability and computational efficiency.

Our experiments demonstrate the dual utility of the simulator as a scalable engine for both policy training and evaluation. We show that imitation policies trained exclusively on simulator-generated data perform comparably to those trained on real-world expert demonstrations. Furthermore, we observe strong correlations between policy performance in the simulator and in the real world across several tasks, validating the Interactive World Simulator as a reliable surrogate for policy training and evaluation.

Looking forward, we plan to extend this framework to more diverse environments and increasingly complex manipulation tasks, further unlocking the potential of large-scale robotic data. An important direction for future work is to study how the performance of world models scales with increasing amounts of interaction data and computational resources. Understanding these scaling behaviors could guide the design of larger and more capable world models for robotics.

\section*{Acknowledgement}

This work was partially supported by the Toyota Research Institute, the DARPA TIAMAT program (HR0011-24-9-0430), NSF Award \#2409661, Samsung Research America, and an Amazon Research Award (Fall 2024). This article solely reflects the opinions and conclusions of its authors and should not be interpreted as necessarily representing the official policies, either expressed or implied, of the sponsors.
We thank Wenlong Huang for valuable suggestions on the project release. We also thank Hongkai Dai, Basile Van Hoorick, Keyi Shen, Zach Witzel, Jaisel Singh, Binghao Huang, Kaifeng Zhang, Hanxiao Jiang, and other RoboPIL members for helpful discussions during the project.

\bibliographystyle{unsrtnat}
\bibliography{references}

\end{document}